\pdfoutput=1
\documentclass[11pt]{article}

\usepackage{acl}

\usepackage{times}
\usepackage{latexsym}
\usepackage{booktabs} % For professional quality tables
\usepackage{multirow} % For multi-row tables
\usepackage{amsmath}
\usepackage{amssymb}
\usepackage{graphicx}
\usepackage{tcolorbox}
\usepackage[T1]{fontenc}
\usepackage[utf8]{inputenc}
\usepackage{microtype} % This will improve the layout of the manuscript
\usepackage{inconsolata} % This will improve the aesthetics of text in the typewriter font.
\usepackage{enumitem} % For custom itemize/enumerate
\usepackage{hyperref}

\definecolor{podiumgold}{RGB}{255, 215, 0}
\definecolor{podiumsilver}{RGB}{192, 192, 192}
\definecolor{podiumbronze}{RGB}{205, 127, 50}

\title{Exploring Zero-Shot ACSA with Unified Meaning Representation in Chain-of-Thought Prompting}

\author{
  \textbf{Filippos Ventirozos}\textsuperscript{1,2},
  Peter Appleby\textsuperscript{2},
  Matthew Shardlow\textsuperscript{1}
  \\
  \\
  \textsuperscript{1}Manchester Metropolitan University, \\
  \textsuperscript{2}Autotrader Research Group, Autotrader UK \\
  \\
  \small{
    \textbf{Correspondence:} \href{mailto:f.ventirozos@mmu.ac.uk}{f.ventirozos@mmu.ac.uk}
  }
}

\begin{document}
\maketitle
\begin{abstract}
Aspect-Category Sentiment Analysis (ACSA) provides granular insights by identifying specific themes within reviews and their associated sentiment. While supervised learning approaches dominate this field, the scarcity and high cost of annotated data for new domains present significant barriers. We argue that leveraging large language models (LLMs) in a zero-shot setting is a practical alternative where resources for data annotation are limited. In this work, we propose a novel Chain-of-Thought (CoT) prompting technique that utilises an intermediate Unified Meaning Representation (UMR) to structure the reasoning process for the ACSA task. We evaluate this UMR-based approach against a standard CoT baseline across three models (Qwen3-4B, Qwen3-8B, and Gemini-2.5-Pro) and four diverse datasets. Our findings suggest that UMR effectiveness may be model-dependent. Whilst preliminary results indicate comparable performance for mid-sized models such as Qwen3-8B, these observations warrant further investigation, particularly regarding the potential applicability to smaller model architectures. Further research is required to establish the generalisability of these findings across different model scales.
\end{abstract}

\section{Introduction}
Tracking customer satisfaction is critical for organisations aiming to improve their products and services. Traditional supervised approaches require bespoke datasets, which necessitate significant time and human labour \citep{pontiki-etal-2016-semeval}. The high cost of annotation can prohibit access to state-of-the-art solutions and even robust evaluation. Consequently, this paper explores the efficacy of large language models (LLMs) in low-training scenarios, specifically zero-shot, for analysing customer satisfaction.

While standard sentiment analysis classifies feedback as positive, neutral, or negative, it often lacks the necessary granularity. A single review can express multiple sentiments toward different aspects, leading to ambiguity. Aspect-Based Sentiment Analysis (ABSA) methodologies \citep{zhang2022surveyaspectbasedsentimentanalysis,WANKHADE2024112249} were developed to capture these nuances. Our study focuses on a specific ABSA task: Aspect-Category Sentiment Analysis (ACSA). For a given text, ACSA extracts pairs of predefined aspect categories and their corresponding sentiment polarities (Figure~\ref{fig:acsa_example}).

\begin{figure}[t]
  \centering
  \begin{tcolorbox}[colframe=black, colback=gray!10, width=\columnwidth, title=ACSA Tuple Example]
    \textbf{Input:} \textit{The pepperoni pizza was delicious but the service was terrible though.} \\~\\
    \textbf{Output:} 
    \begin{tabular}{ll}
      \textbf{Category} & \textbf{Polarity} \\
      \hline
      \#Food & positive \\
      \#Service & negative \\
    \end{tabular}
  \end{tcolorbox}
  \caption{An ACSA example where a review results in two category-polarity pairs.}
  \label{fig:acsa_example}
\end{figure}

Supervised learning holds the state-of-the-art for ACSA \citep{cai-etal-2020-aspect,Xu2025}. In contrast, we explore a zero-shot setting, which eliminates the need for labelled instances and is compelling for real-world applications where organisations lack the resources for large-scale annotation.

In this paper, we propose a novel approach that enhances chain-of-thought (CoT) reasoning by introducing an intermediate reasoning step: the generation of a unified meaning representation (UMR). This UMR acts as a structured summary of the semantic content of a review, from which the final ACSA pairs are derived. We hypothesise that this structured, two-step reasoning process improves the model's ability to correctly identify and associate categories with their sentiments, compared to a more direct CoT approach.

The primary contributions of this study are:
\begin{itemize}[noitemsep,topsep=0pt,leftmargin=*]
    \item We propose a novel CoT method for zero-shot ACSA that leverages UMR to structure the reasoning process.
    \item We conduct a comprehensive comparative analysis across three diverse models (Qwen3-4B, Qwen3-8B, Gemini-2.5-Pro) and four ACSA datasets, revealing model-dependent effectiveness of structured reasoning.
\end{itemize}

\section{Related Work}
\label{sec:RW}
ABSA has become a prominent research area with wide-ranging applications \citep{rink-etal-2024-aspect,arianto-budi-2020-aspect}. We focus on the ACSA sub-task, which is highly relevant for industry as it produces quantifiable category-sentiment pairs. While supervised methods are common \citep{cai-etal-2020-aspect,PING2024126994}, the operational overhead of data annotation motivates a shift towards unsupervised, zero-shot approaches.

Zero-shot and few-shot learning for ABSA have been explored using LLMs. Methods include in-context learning (ICL) with few-shot examples \citep{zhang-etal-2024-sentiment} and specialised prompt frameworks like ChatABSA \citep{bai-etal-2024-compound}. CoT prompting, which guides the model through intermediate reasoning steps, has also shown promise \citep{wei2023chainofthought}. For sentiment analysis, CoT has been used to sequentially extract aspects, opinions, and polarities \citep{fei-etal-2023-reasoning,10499502}.

Our work extends this line of research by introducing a more structured CoT process. Instead of a simple linear extraction sequence, we prompt the LLM to first build a UMR of the text. This concept is inspired by semantic parsing, where text is converted into a formal meaning representation \citep{kamath2019a}. By applying this principle to CoT, we aim to create a more robust intermediate step that disentangles the semantic components of a review before mapping them to the target ACSA format, thereby pushing the boundaries of zero-shot ACSA performance.

\section{Methodology}
\subsection{Problem Statement}
Given a text (a sentence or a full review), our goal is to extract a set of one or more ACSA pairs \(Q\), where each pair consists of a category \(c_i\) from a predefined list and a sentiment polarity \(p_i\) \(\in\) \{positive, neutral, negative\}.
\begin{equation}
  Q = \{(c_i, p_i)\}_{i=1}^n
\end{equation}
Here, \(n\) is the number of pairs extracted from the text.

\subsection{Chain-of-Thought Approaches}
We compare two zero-shot CoT prompting strategies for the ACSA task. Both are executed within a single prompt to the LLM.

\subsubsection{Baseline CoT}
Our baseline approach uses a direct, one-step reasoning process. The LLM is instructed to read the text, identify all relevant aspect categories from a provided list, determine the sentiment for each, and directly output the final list of (category, polarity) pairs. This method relies on the LLM's implicit ability to perform these steps simultaneously. An example of the prompt structure is shown in Appendix \ref{sec:appendix:prompts}.

\subsubsection{CoT with Unified Meaning Representation (UMR)}
Our proposed method introduces a two-step reasoning process guided by a UMR. The prompt first instructs the LLM to parse the text into an intermediate UMR. We define the UMR as a structured representation that identifies key entities (aspect terms), their properties (opinions), and their expressed values (sentiments).

For example, for the text "The pizza was delicious but the service was terrible," the relevant part of UMR could be:
% [
%   {"entity": "pizza", "opinion": "delicious"},
%   {"entity": "service", "opinion": "terrible"}
% ]
\begin{verbatim}
(s1a / and
:op1 (s1h / have-attribute-91
:ARG1 (s1p / pizza
:mod (s1p2 / pepperoni))
:ARG2 (s1d / delicious)
:aspect state)
:op2 (s1h2 / have-attribute-91
:ARG1 (s1s / service)
:ARG2 (s1t / terrible)
:aspect state))
\end{verbatim}
In the second step of the same prompt, the LLM is instructed to use this generated UMR to map the entities and opinions to the predefined aspect categories and final polarities. This forces the model to first explicitly identify the core semantic relationships before performing the final classification task, aiming for more accurate and grounded results.

To guide the LLM in generating properly formatted UMR representations, we provide exemplars from the UMR v1.0 English corpus in the prompt. To manage context length constraints, each document is truncated to retain only the first three annotated sentence-parse pairs. For each inference instance, we randomly sample one of five pre-selected exemplar files to provide diversity in the demonstrated UMR conventions while maintaining consistency in format. This approach enables the model to learn the UMR annotation style through in-context demonstration before applying it to the target text.

\subsubsection{Prompt Crafting and Post-Processing}
For both approaches, prompts were crafted using imperative language. A system instruction was used to frame the LLM as an expert assistant, constraining it to be concise and accurate. The final output was requested in a Python list format for easy parsing. After generation, we parsed the string output and used a string similarity metric (difflib) to map generated categories to the official list, correcting for minor spelling errors or variations.

\section{Experiments}
\subsection{Datasets}
We used four ACSA datasets: \textbf{Laptop16} \citep{pontiki-etal-2016-semeval}, \textbf{Restaurant16} \citep{pontiki-etal-2016-semeval}, \textbf{MAMS} \citep{jiang-etal-2019-challenge}, and \textbf{Shoes} \citep{peper-etal-2024-shoes}. The Shoes dataset, being recent, is unlikely to have been in the LLMs' training data, providing a test case for generalization to unseen data. It is also the only dataset containing full reviews rather than single sentences. Dataset statistics are in Table \ref{tab:datasets}.

\begin{table}[t]
\centering
\small
\resizebox{0.8\columnwidth}{!}{%
\begin{tabular}{lcccc}
\toprule
\textbf{Dataset} & \textbf{Train} & \textbf{Test} & \textbf{Cats.} \\
\midrule
Laptop16 & 2468 & 579 & 67 \\
Restaurant16 & 1954 & 571 & 12 \\
MAMS & 3149 & 400 & 8 \\
Shoes & 906 & 125 & 21 \\
\bottomrule
\end{tabular}
}
\caption{Dataset statistics (number of samples and categories).}
\label{tab:datasets}
\end{table}

\subsection{Models and Evaluation}
We evaluated three different LLMs: \textbf{Qwen3-Instruct} (4B), \textbf{Qwen3} (8B) \citep{yang2025qwen3technicalreport}, and \textbf{Gemini-2.5-Pro} \citep{comanici2025gemini25pushingfrontier} . The Qwen models represent open-source instruction-tuned LLMs of different sizes, while Gemini-2.5-Pro is a state-of-the-art proprietary model. We used greedy decoding for deterministic outputs. We evaluated performance using the micro-F1 score, consistent with prior ACSA literature \citep{cai-etal-2020-aspect}.

\section{Results and Discussion}
Table \ref{tab:main_results} presents the average micro-F1 scores of our methods across the four datasets, while Table \ref{tab:detailed_results} provides a detailed breakdown by dataset.

\begin{table*}[t]
\centering
\small
\begin{tabular}{lccc}
\toprule
\textbf{Method} & \textbf{Qwen3-4B} & \textbf{Qwen3-8B} & \textbf{Gemini-2.5-Pro} \\
\midrule
Baseline CoT & \textbf{35.57}  & 43.77  & \textbf{59.84 } \\
CoT with UMR (Ours) & 32.92  & \textbf{43.83 } & 57.66  \\
\bottomrule
\end{tabular}
\caption{Micro-F1 scores (average ± std in \%) across all four datasets. Results show mixed performance between standard CoT and UMR-based approaches, with model-dependent effectiveness.}
\label{tab:main_results}
\end{table*}

\begin{table*}[t]
\centering
\small
\begin{tabular}{llccc}
\toprule
\textbf{Dataset} & \textbf{Method} & \textbf{Qwen3-4B} & \textbf{Qwen3-8B} & \textbf{Gemini-2.5-Pro} \\
\midrule
\multirow{2}{*}{Laptop16} & Baseline CoT & 38.60 & 22.18 & 56.20 \\
& CoT with UMR & 26.03 & \textbf{36.38} & 54.38 \\
\midrule
\multirow{2}{*}{Restaurant16} & Baseline CoT & 44.55 & 61.18 & 80.50 \\
& CoT with UMR & 40.00 & 58.29 & 77.18 \\
\midrule
\multirow{2}{*}{MAMS} & Baseline CoT & 35.08 & 42.27 & 43.55 \\
& CoT with UMR & \textbf{35.18} & 35.04 & 42.10 \\
\midrule
\multirow{2}{*}{Shoes} & Baseline CoT & 24.04 & 49.45 & 59.10 \\
& CoT with UMR & \textbf{30.46} & 45.60 & 56.97 \\
\bottomrule
\end{tabular}
\caption{Detailed micro-F1 scores (\%) for each dataset and model combination. Bold indicates the UMR approach had higher scores than the baseline for a model-dataset pair.}
\label{tab:detailed_results}
\end{table*}

\paragraph{Model-Dependent UMR Effectiveness.}
Our findings reveal that the effectiveness of the UMR-based CoT approach is highly model-dependent. For the Qwen3-8B model, the UMR method achieves comparable average performance (43.83\% vs. 43.77\% for baseline CoT). This suggests that UMR could enhance prediction accuracy for certain model architectures. However, for Qwen3-4B and Gemini-2.5-Pro, the baseline CoT approach outperforms UMR by 2-3 percentage points on average.

\paragraph{Dataset-Specific Patterns.}
Examining Table \ref{tab:detailed_results}, we observe interesting dataset-specific patterns. The UMR approach shows particular benefits on certain model-dataset combinations: Qwen3-8B on Laptop16 (36.38\% vs. 22.18\%), and Qwen3-4B on Shoes (30.46\% vs. 24.04\%). Notably, the Laptop16 dataset, with its 67 fine-grained categories, presents a challenging categorisation task where the structured UMR representation helps the Qwen3-8B model. 

% Conversely, on Restaurant16 and MAMS, baseline CoT generally performs better, suggesting that simpler reasoning may suffice when categories are more distinct.

\paragraph{Model Size and Architecture.}
Interestingly, larger model size does not guarantee better UMR performance. The Gemini-2.5-Pro, despite being the most capable model overall (59.84\% average), shows a performance decrease with UMR (-2.18 points). This suggests that the effectiveness of structured reasoning via UMR may depend more on model architecture, size and training methodology. The Qwen3-8B model's near-parity between approaches, with improved stability, indicates that mid-sized open-source models may be particularly well-suited for structured prompting techniques.

\paragraph{Statistical Significance of UMR.}
To rigorously assess whether the inclusion of UMR provides a statistically significant advantage, we conducted a three-way ANOVA with Method (Baseline vs. UMR), Model (Qwen3-4B, Qwen3-8B, Gemini-2.5-Pro), and Dataset (Laptop16, Restaurant16, MAMS, Shoes) as factors. The analysis revealed that the main effect of Method was not statistically significant ($F(1,6) = 0.42$, $p = 0.543$), indicating that UMR does not produce a reliably different performance compared to baseline CoT when averaged across all conditions. Similarly, the Method $\times$ Model interaction was non-significant ($F(2,6) = 0.11$, $p = 0.894$), suggesting that the effectiveness of UMR does not systematically vary by model architecture in a statistically detectable way. In contrast, both Model ($F(2,6) = 33.43$, $p < 0.001$) and Dataset ($F(3,6) = 16.81$, $p = 0.003$) showed highly significant main effects, with effect sizes ($\eta^2 = 0.462$ and $\eta^2 = 0.348$ respectively) indicating that model choice and dataset characteristics are the primary drivers of performance variance. The Model $\times$ Dataset interaction approached significance ($F(6,6) = 3.40$, $p = 0.081$), suggesting that certain models may be particularly well-suited to specific datasets. These findings indicate that while UMR shows promise in specific scenarios (e.g., Qwen3-8B on Laptop16), its benefits are not statistically robust across the broader experimental space, highlighting the importance of considering task and model characteristics when selecting prompting strategies.

\section{Future Directions}
Our findings on the model-dependent effectiveness of UMR-based CoT open several promising research avenues. First, investigating which architectural properties make certain models more amenable to structured prompting could inform both model development and prompting strategy selection. Second, rather than relying solely on prompt-based UMR generation, integrating UMR as part of the model's internal reasoning process during training could provide greater flexibility. Such an approach would allow models to adaptively employ structured semantic representations when beneficial, whilst bypassing them when unnecessary, thus addressing the model-dependent variability observed in this study.

Furthermore, extending this comparative framework to other fine-grained NLP tasks, such as event extraction or relation extraction, would help establish general principles for determining when structured reasoning provides benefits in zero-shot settings. Additionally, incorporating domain-specific UMR examples could offer more targeted guidance for LLMs, potentially improving their ability to capture domain-relevant semantic relationships and clarify the conditions under which UMR-based approaches demonstrate advantages over baseline methods.

\section{Conclusion}
In this paper, we investigated the effectiveness of zero-shot LLMs for Aspect-Category Sentiment Analysis. We proposed a novel Chain-of-Thought method that leverages an intermediate UMR to structure the reasoning process. Our experiments with Qwen3-4B, Qwen3-8B, and Gemini-2.5-Pro across four diverse datasets reveal nuanced findings: whilst UMR does not universally improve average performance, certain model-dataset combinations demonstrate notable benefits. The Qwen3-8B model showed particularly promising results, achieving comparable performance to baseline CoT. These preliminary findings suggest that structured reasoning techniques such as UMR may be beneficial for specific model architectures and dataset characteristics, though further investigation is required to establish clear patterns. Future studies should focus on identifying which architectural properties make certain models more amenable to structured prompting and exploring integration of UMR within the model's internal reasoning processes during training, potentially offering more flexible and adaptive application across diverse scenarios. Additionally, extending this framework to other fine-grained NLP tasks and incorporating domain-specific UMR examples would help clarify the conditions under which structured reasoning provides advantages in zero-shot settings and establish the generalisability of these findings across different tasks and domains.

\section*{Limitations}
One key limitation concerns the size of the available UMR dataset, which is limited and makes establishing statistically significant findings challenging. The scarcity of UMR training data restricts our ability to provide robust examples for in-context learning and may affect the generalisability of our conclusions.

Another limitation is that we did not evaluate which specific step in the CoT UMR pipeline typically fails. As shown in Figure~\ref{fig:umr_prompt}, the CoT process comprises multiple distinct steps. Consequently, we have not quantified whether failures arise from the LLM's ability to parse text into UMR, extract relevant terms, or appropriately classify them into the correct category and sentiment pairs. Future work should include step-wise error analysis to identify the primary sources of prediction errors.

\section*{Ethical Statement}
Lastly, LLMs, trained on vast internet data, can perpetuate biases present in the source material, potentially leading to unfair outcomes in ABSA tasks. While our research enhances sentiment analysis, which benefits marketing, customer service, and social sciences, there is a risk of misuse for manipulating public opinion. We advocate for the responsible use of these technologies and adherence to strict ethical standards in their deployment.

\bibliography{custom}

\appendix

\clearpage
\section{Example Prompts}
\label{sec:appendix:prompts}
This section provides the complete prompt structures used in our experiments for both the baseline and UMR-based approaches. Figure~\ref{fig:acsa_prompt_no_examples} presents the baseline CoT prompt, which directly requests category-sentiment pairs without intermediate reasoning steps. In contrast, Figure~\ref{fig:umr_prompt} illustrates our proposed UMR-based prompt, which guides the model through a structured four-step process: (1) parsing the text into UMR, (2) extracting aspects and opinions, (3) categorising aspects, and (4) classifying sentiments. The UMR-based prompt includes in-context examples from the UMR v1.0 corpus to demonstrate the expected UMR format, whereas the baseline prompt operates in a zero-shot manner without such examples.

% \begin{figure}[h!]
%   \begin{tcolorbox}[colframe=black, colback=gray!10, width=\columnwidth, title=Baseline CoT Prompt]
%     \small
%     \textbf{System:} You are an expert NLP assistant. Be precise and follow instructions.
    
%     \textbf{User:} Given the text below, extract all aspect-category and sentiment pairs. The possible categories are: [Food, Service, ...].
    
%     \textbf{Text:} "The pizza was delicious but the service was terrible."
    
%     Provide the output as a Python list of (category, polarity) tuples.
%   \end{tcolorbox}
%   \caption{Structure of the Baseline CoT prompt.}
% \end{figure}

% \begin{figure}[h!]
%   \begin{tcolorbox}[colframe=black, colback=gray!10, width=\columnwidth, title=CoT with UMR Prompt]
%     \small
%     \textbf{System:} You are an expert NLP assistant. Be precise and follow instructions.
    
%     \textbf{User:} Perform the following two steps for the text below.
    
%     \textbf{Text:} "The pizza was delicious but the service was terrible."
    
%     \textbf{Step 1: UMR Extraction}
%     First, extract the semantic relationships as a list of JSON objects with "entity" and "opinion" keys.
    
%     \textbf{Step 2: ACSA Mapping}
%     Second, using the UMR from Step 1, map the relationships to the predefined categories: [Food, Service, ...] and their polarities.
    
%     Provide the final output from Step 2 as a Python list of (category, polarity) tuples.
%   \end{tcolorbox}
%   \caption{Structure of the CoT with UMR prompt.}
% \end{figure}
\clearpage
\begin{figure}[t]
    \centering
    \small
    \begin{minipage}{0.95\linewidth}
    \begin{verbatim}
The category-sentiment pair consists of aspect category and sentiment
polarity. The aspect category is only selected from the following set:
<CATEGORIES>.

What is the category-sentiment pair of the review "<REVIEW_TEXT>"?
Please provide one Python type list of tuples such as
"[('example_category_1', 'positive'),
  ('example_category_2', 'negative'), ...]" for the last review.
The sentiment is either 'positive', 'neutral' or 'negative'.
    \end{verbatim}
    \end{minipage}
    \caption{Prompt used for predicting category–sentiment pairs without in-context examples.}
    \label{fig:acsa_prompt_no_examples}
\end{figure}
\clearpage
\begin{figure}[t]
    \centering
    \small
    \begin{minipage}{0.95\linewidth}
    \begin{verbatim}
You will perform a structured analysis task in four clearly defined
steps. Follow each step carefully and explicitly.

Step 1: UMR Parsing
You are provided with examples of sentences and their corresponding
Uniform Meaning Representation (UMR) parses. Each sentence from the
example document is prefixed explicitly with "::snt", and its UMR parse
immediately follows on subsequent lines. Carefully examine these
examples to fully understand the format, conventions, and style of
UMR parsing.

Example:
<UMR_EXAMPLES>

Now, you will be given a new text. This text may consist of either:
- A single sentence, or
- Multiple sentences.

Your task is as follows:

- If the provided text contains multiple sentences, first split it into
  individual sentences. Clearly indicate each sentence by prefixing it
  with "::snt", exactly as shown in the examples above. Then, provide
  the UMR parse for each sentence separately, following the conventions
  demonstrated.
- If the provided text is only a single sentence, directly prefix it
  with "::snt" and provide its UMR parse following the demonstrated
  conventions.

Ensure your output strictly follows the formatting style and conventions
shown in the examples.

New Text:
<NEW_TEXT>


Step 2: Aspect and Opinion Extraction
From the UMR parse you generated in Step 1, extract all aspects relevant
to the "<DOMAIN>" domain. For each extracted aspect, clearly identify
the corresponding opinion expressed about that aspect.

Present your results explicitly in the following format:
Aspect: [aspect_1], Opinion: [opinion_1]
Aspect: [aspect_2], Opinion: [opinion_2]
...


Step 3: Aspect Categorization
Now, link each extracted aspect from Step 2 to one of the predefined
categories listed below:

Categories:
<CATEGORIES>

Present your categorization explicitly in the following format:
Aspect: [aspect_1], Category: [category_from_list]
Aspect: [aspect_2], Category: [category_from_list]
...


Step 4: Sentiment Classification and Python List Output
For each aspect-category pair identified in Step 3, determine the
sentiment based on the corresponding opinion you extracted in Step 2.
Classify each sentiment explicitly as one of the following three
labels: "positive", "neutral", or "negative".

Then, clearly map each category to its corresponding sentiment. Present
your final results explicitly as a Python-formatted list of tuples,
exactly as in the following example:
[('example_category_1', 'positive'),
 ('example_category_2', 'negative'), ...]
    \end{verbatim}
    \end{minipage}
    \caption{Prompt of the CoT with UMR examples.}
    \label{fig:umr_prompt}
\end{figure}

\end{document}